\documentclass{article}
\usepackage{spconf,amsmath,amssymb,graphicx}
\usepackage[pagebackref=true,breaklinks=true,letterpaper=true,colorlinks,bookmarks=false]{hyperref}

\title{Speech Prediction in Silent Videos using Variational Autoencoders}
\name{Ravindra Yadav$^{\star}$ \qquad Ashish Sardana$^{\dagger}$ \qquad Vinay P Namboodiri$^{\ddagger}$ \qquad Rajesh M Hegde$^{\star}$}
\address{$^{\star}$ Indian Institute of Technology Kanpur, India \\
$^{\dagger}$ NVIDIA\\
$^{\ddagger}$ University of Bath, UK}

\begin{document}
%
\maketitle
\begin{abstract}
Understanding the relationship between the auditory and visual signals is crucial for many different applications ranging from computer-generated imagery (CGI) and video editing automation to assisting people with hearing or visual impairments. However, this is challenging since the distribution of both audio and visual modality is inherently multimodal. Therefore, most of the existing methods ignore the multimodal aspect and assume that there only exists a deterministic one-to-one mapping between the two modalities. It can lead to low-quality predictions as the model collapses to optimizing the average behavior rather than learning the full data distributions. In this paper, we present a stochastic model for generating speech in a silent video. The proposed model combines recurrent neural networks and variational deep generative models to learn the auditory signal's conditional distribution given the visual signal. We demonstrate the performance of our model on the GRID dataset based on standard benchmarks.
\end{abstract}
\section{Introduction}
\label{sec:intro}
Real-world observations often involve multiple different modalities that are synchronized with each other. Thus, the same underlying phenomena can be described using a multitude of different perspectives. For example, a concept dog can be described by an image, a text (caption), and an audio clip (barking sound). Thus, learning the correspondence between the different modalities can help in various real applications. The cross-modal generation has, therefore, become an increasingly important problem in recent years. However, building an artificial agent that can learn the underlying correlation between statistically different modalities has turned out to be extremely challenging than expected.

In this work, we are interested in learning correspondence between visual and auditory signals for speech prediction in silent videos. The ability to predict auditory information given visual signal is useful for many real applications, such as assisting people with hearing impairments, dubbing movies, video editing automation, and enhancing virtual reality systems. However, due to the high dimensionality of audio and video streams and spatio-temporal complexities in natural videos, predicting the audio modality from raw sensory observations such as videos is exceptionally difficult.

Prior approaches for speech prediction in silent videos, therefore, make various simplifying assumptions. One particularly common assumption is that there exists a one-to-one mapping between the audio and visual modalities; thus, a deterministic model will suffice. However, a deterministic model cannot capture the nuances that are present in real-world videos. As a result, a deterministic model ends up optimizing the average behavior with no notion of uncertainty (or diversity). Thus, a stochastic modeling approach is needed that can learn the full data distributions of the given modalities in a principled way.

\section{Related Work}
Recent developments in deep learning have led to impressive results in speech generation and prediction. Text-to-speech systems \cite{Char2Wav,Tacotron}, which uses text or phonemes as inputs to generate audio, have been widely useful in ubiquitous applications. However, there are many applications where the input is a video instead of a text input. In those cases, incorporating a TTS system as an intermediate encoding network requires non-trivial network modifications, which introduces new training challenges. Thus, there has been a recent surge in designing deep learning models that can directly map the frame stream to the audio stream in a given video.

Using hand-crafted features, systems for producing audio speech from only visual features are proposed in \cite{vid2speech_handcrafted1,vid2speech_handcrafted2}. Ariel et. al. \cite{Vid2speech} proposed an end-to-end trainable approach that predicts the LPC (Linear Predictive Coding) features given a set of video frames. The approach uses a clip of K consecutive video frames as an input to a 2D-CNN network that is trained to predict the LPC features, these features are then converted into intelligible speech. However, due to the lack of any recurrent units, the generated speech is not temporally coherent, thus lack naturalness. Later in \cite{Vid2speech_improved}, the same research group proposed using mel spectrograms features, instead of the LPC features. In addition, the encoder module was modified to also take into account the optical flow information computed between frames.

To capture temporal information, in \cite{Lip2audspec}, Hassan et. al. incorporated Long Short-Term Memory (LSTM) \cite{lstm} units to encode the visual information. The encoded features are then passed to a pre-trained autoencoder network to predict the audio features. Using an adversarial training approach, in \cite{vid2speech_vougioukas}, the author proposed using a discriminator network (called "critic") that distinguishes real audio from the generated audio. However, the model uses a pre-trained speech-to-video network \cite{speech2vid_vougioukas} as a reference model to minimize the perceptual loss between the features obtained from real and generated audios. Recently, Prajwal et. al. \cite{lip2wav} proposed a Tacotron2 model \cite{Tacotron2} variant for visual to speech generation. The model consists of a stack of 3D convolution layers that encode the frames' sequence into a fixed-size feature vector. The obtained feature vector is then decoded by the attention-based recurrent neural network to generate the audio signal's mel spectrogram features.

In contrast to the earlier approaches that follow a discriminative modeling approach, we propose a generative approach that learns the distribution of speech signal given the video frames in the following sections. The introduction of the latent variables into the modeling allows the model to capture every nuance present in the natural speech signal.

\section{The Proposed Model}
Suppose we are given a video that consists of a frame stream $\mathbb{F}$=\{$f_1$, $f_2$, ..., $f_N$\}, and audio stream $\mathbb{A}$=\{$a_1$, $a_2$, ..., $a_N$\}, such that both the sets have equal number of elements N. In our model we preprocess the audio so that the elements $a_1, a_2, ..., a_N$ represent the mel spectrograms features, instead of the raw audio. Further, rather than using individual frames $f_i$ as an input to the model, we use a short clip of K consecutive frames (we use $K=5$). The clip contains frame $f_i$ at the center and a few of its neighboring frames, which serves as a context. Using context frames has been shown to improve the accuracy in speech prediction tasks, and therefore is used in almost every deep learning model discussed above.

\begin{figure}[t]
\begin{minipage}[b]{1.0\linewidth}
  \centering
  \centerline{\includegraphics[width=\textwidth]{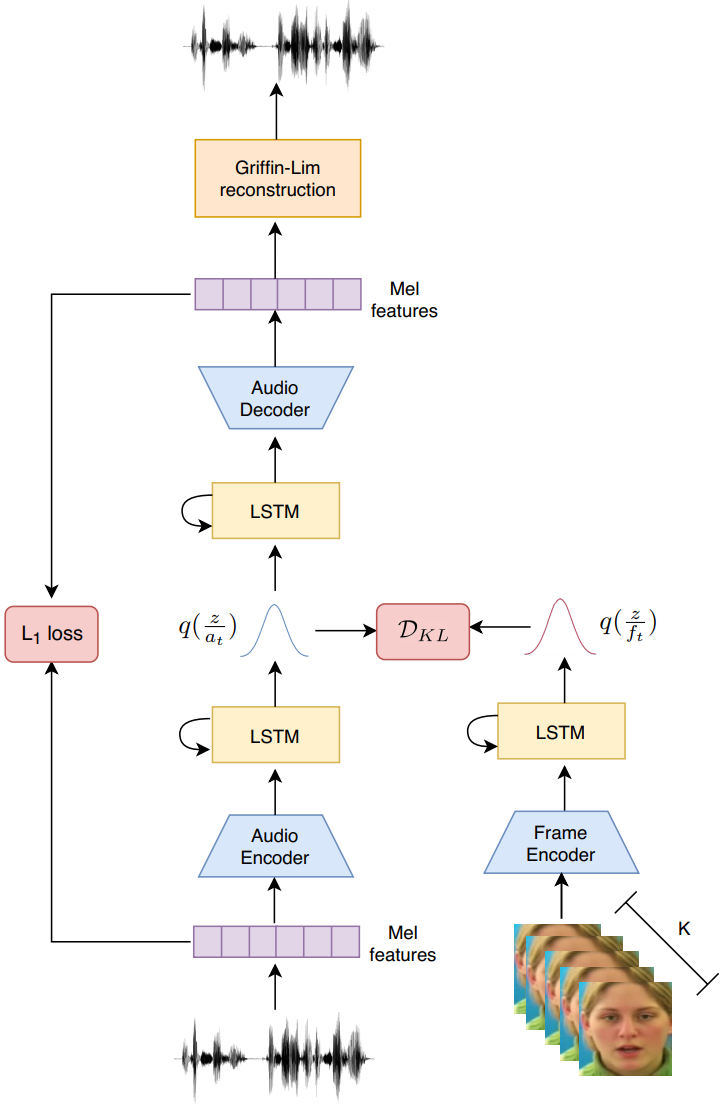}}
\end{minipage}
\caption{Overview of proposed speech prediction model.}
\label{proposed_model}
\end{figure}

Our objective is to learn a mapping function H such that H($f_i$)=$a_i$, where $i$ represent any $i_{th}$ element of the two sets. We could learn H as a simple regression function that independently maps each element $a_i$ to the corresponding element $f_i$. It, however, could result in an output that has temporal discontinuities. Therefore, to generate any $a_i$ element, we also take into account the previously generated $a_i$'s, such that H($f_i$, $a_{i-1}$, $a_{i-2}$, $a_1$) = $a_i$ is an autoregressive function. In the proposed model, we model function H using a combination of variational autoencoders and recurrent neural networks so that, at every time step, we can capture the distribution of audio speech using a latent variable $z$.

Variational autoencoders \cite{VAEKingma,VAERezende} are class of deep generative models used for unsupervised learning. Given a dataset $X = \{x_1, x_2, ... x_N\}$, we typically aim at maximizing the marginal log-likelihood $\text{ln} p(X)\text{=}\sum\limits_{i=1}^N \text{ln} p(x_i)$. However, when using neural networks, the direct optimization of the marginal likelihood $p(X)$ is computationally intractable. To solve this issue VAEs uses an approximate inference model $q_\phi (\frac{z}{x})$ that serves as an approximation to the true posterior distribution $p_\theta (\frac{z}{x})$. During training, we jointly maximize the below variational lower bound to the marginal log-likelihood with respect to parameters $\theta$ and $\phi$,

\begin{equation}
\begin{array}{r@{}l}
\begin{aligned}
\mathcal{L(\theta, \phi; \textnormal{x})} =  \mathbb{E}_{q_{\phi}(z|x)}[\textup{log}\hspace{0.05cm} p_\theta(x|z)] - \mathcal{D}(q_{\phi}(z|x) ||  p_\theta(z))
\end{aligned}
\end{array}
\end{equation}

where $\mathcal{D}$ denotes the Kullback-Leibler divergence, $\mathcal{L}$ is the evidence lower bound (abbr. ELBO).\\

\textbf{VAEs for Speech Prediction.} For the speech prediction in silent videos, we adapt the above VAE formulation so that it takes into account both audio and frame modalities. Thus, in the proposed model, the objective we maximize is given by,
\vspace{-0.1in}
\begin{equation}
\begin{array}{r@{}l}
\begin{aligned}
\mathcal{L(\theta, \phi; \textnormal{x})} &= \sum\limits_{t=1}^N \mathbb{E}_{q_{\phi_{a}}(z|a_t)}[\lambda\textup{log}\hspace{0.05cm} p_{\theta_{a}}(a_t|z)] \\
&- \beta KL[q_{\phi_{a}}(z|a_t) || q_{\phi_{f}}(z|f_t)]
\end{aligned}
\end{array}
\label{proposed_equation}
\end{equation}

In above equation, we model both audio and frame posterior distributions using a Gaussian distribution with diagonal covariances i.e. $q_\phi(z|\cdot) = \mathcal{N}(z|\mu_\phi(\cdot),diag(\sigma_{\phi}^{2}(\cdot)))$. In our experiments, based on the performance on validation data set, we set $\lambda$ and $\beta$ values equal to 1 and 1e-6 respectively.

Figure~\ref{proposed_model} shows a higher level pictorial representation of our model. The model consists of an autoencoder module that encodes the audio stream using a latent distribution $q(z|a)$; additionally, there is an encoder network for the frame stream whose output serves as a prior distribution for the audio autoencoder module.

\textbf{Architectures.} We first embed both audio and frame modalities into fixed-size feature vectors using different encoder networks. The audio encoder network is modeled as a 3-layered fully connected network, while the frame encoder network uses the same architecture as VGG16 \cite{vgg16}. These obtained feature vectors then passed through two different LSTMs to learn the temporal information present in the audio and frame streams. For the audio stream, it introduces the following recurrence,

\begin{equation}
\begin{array}{r@{}l}
\begin{aligned}
e_a^t &= \phi(a_t; W_{ee}) \\
h_{a}^t &=LSTM(h_{a}^{t-1}, e_a^t; W_{encoder})
\end{aligned}
\end{array}
\label{eit}
\end{equation}

where $W_{ee}$ and $W_{encoder}$ are the embedding weights and the LSTM cell weights, respectively. The feature vector $h_{a}^t$ is further passed through two different multilayer perceptrons, $\delta_1$ and $\delta_2$, to obtain the mean and variance (we basically obtain log of variance) parameters for the posterior distribution $q(z|a_t)$.

\begin{equation}
\begin{array}{r@{}l}
\begin{aligned}
\mu_a &= \delta_1(h_{a}^t; W_{\delta_1}) \\
\sigma_a^2 &= \delta_2(h_{a}^t; W_{\delta_2})
\end{aligned}
\end{array}
\end{equation}

where $W_{\delta_1}$ and $W_{\delta_2}$ are embedding weights of the respective MLPs. Similarly, for the frame stream, at every time step, we obtain the parameters $\phi = (\mu_f, \sigma_f^2)$ for the Gaussian distribution $q_\phi(z|f_t) = \mathcal{N}(z|\mu_f, diag(\sigma_f^2))$.

At the decoder end, the sample $z_t$ from the posterior distribution $q(z|a_t)$ is passed through another LSTM, the output of which is then fed to an audio decoder network that is a mirrored version of the audio encoder network. This gives us the mel spectrogram features, which are then used to reconstruct the time-domain audio signal using the Griffin-Lim algorithm.

\textbf{Speech prediction at test time.} At test time, since we are only given the frame stream $\mathbb{F}$=\{$f_1$, $f_2$, ..., $f_N$\}. Thus, at every time-step t, we obtain the posterior distribution $q(z|f_t)$ using the frame encoder network. A sample from the obtained posterior distribution is then fed to the audio decoder network to generate the audio waveform.

\section{Results}
In this section, we present the results on the publicly available GRID dataset \cite{gridcorpus}. The GRID dataset consists of videos of 33 speakers, each uttering 1000 different sentences. Each sentence contains six words chosen from a fixed dictionary. Similar to prior works \cite{Lip2audspec,vid2speech_vougioukas,lip2wav}, we used four speakers (S1, S2, S4, and S29) from the dataset for comparison.

In the following sections, we show quantitative comparisons based on the Short-Time Objective Intelligibility (STOI) \cite{stoi} and Extended Short-Time Objective Intelligibility (ESTOI) \cite{estoi} and Perceptual Evaluation of Speech Quality (PESQ) \cite{pesq} metrics. We also make a qualitative comparison between samples from our model and current state-of-the-art Lip2Wav model \cite{lip2wav}. As mentioned before, speech prediction is a many-to-many mapping problem. Therefore, we evaluate our model based on the diversity aspect that have previously not all been considered by one model. We show that our proposed model can generate multiple different plausible audio speech given the same input video.

\begin{figure*}[]
\includegraphics[width=\textwidth]{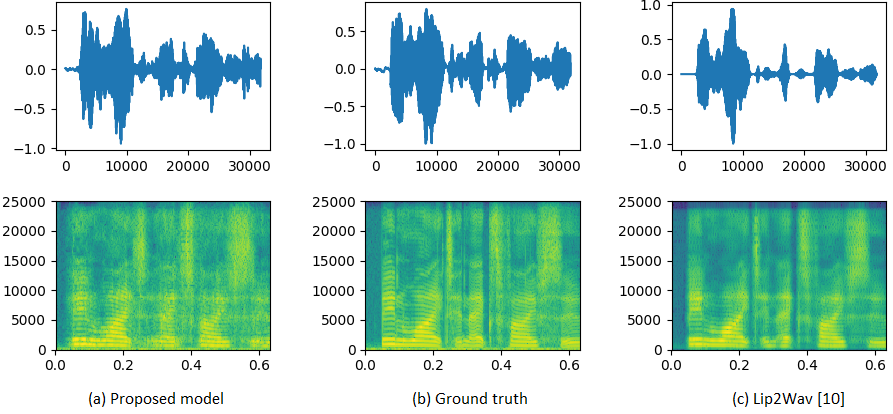}
\caption{A qualitative comparison between the proposed model and state-of-the-art Lip2Wav \cite{lip2wav} model.}
\vspace{-0.1in}
\label{qualitative_waveform_and_spectrogram}
\end{figure*}

\subsection{Quantitative Evaluation}
We compare our method on three metrics SOTI, ESTOI, and PESQ, against different baselines in Table~\ref{quantitative_sota}.

\begin{table}[h]
\begin{center}
\caption{Quantitative evaluation: A comparison between mean STOI, ESTOI and PESQ scores for different models.}
\vspace{0.1in}
\label{quantitative_sota}
\renewcommand{\arraystretch}{1.2}
\begin{tabular}{|c|c|c|c|}
\hline
Model & STOI & ESTOI & PESQ  \\
\hline
Vid2Speech [5] & 0.491 & 0.335 & 1.734 \\
\hline
Lip2AudSpec [7] & 0.513 & 0.352 & 1.673 \\
\hline
Konstantinos et. al. [8] & 0.564 & 0.361 & 1.684 \\
\hline
Ephrat et al. [6] & 0.659 & 0.376 & 1.825 \\
\hline
Lip2Wav [10] & \textbf{0.731} & 0.535 & 1.772 \\
\hline
Proposed model & 0.724 & \textbf{0.540} & \textbf{1.932} \\
\hline
\end{tabular}
\end{center}
\end{table}
We found that models that use spectrogram features perform much better than the trained models using raw audio samples or Line Spectrum Pairs (LSPs). This might be due to difficulty in reconstructing the high dimensional raw audio, sampled at 16KHz, using the standard L1 (or L2) loss functions. On the other hand, due to the absence of original excitation, speech generated using LSP features are unnatural and robotic. Our probabilistic model, which can capture human speech nuances, outperforms the other existing models that also use spectrogram features for training. This shows the advantage of stochastic modeling that takes into account crucial aspects such as variance and multimodality.

\subsection{Qualitative Evaluation}
Figure~\ref{qualitative_waveform_and_spectrogram} shows a qualitative comparison between our proposed model and currently state-of-the-art Lip2Wav model \cite{lip2wav}. We found that the Lip2Wav model is biased towards overly smooth results that might explain the lower PESQ score, which measures speech quality, obtained by the Lip2Wav model. On the other hand, our proposed model produces much better quality audio that closely matches the ground truth.

\subsection{Diverse Predictions}
\begin{figure}[t]
\begin{minipage}[b]{1.0\linewidth}
  \centering
  \centerline{\includegraphics[width=\textwidth]{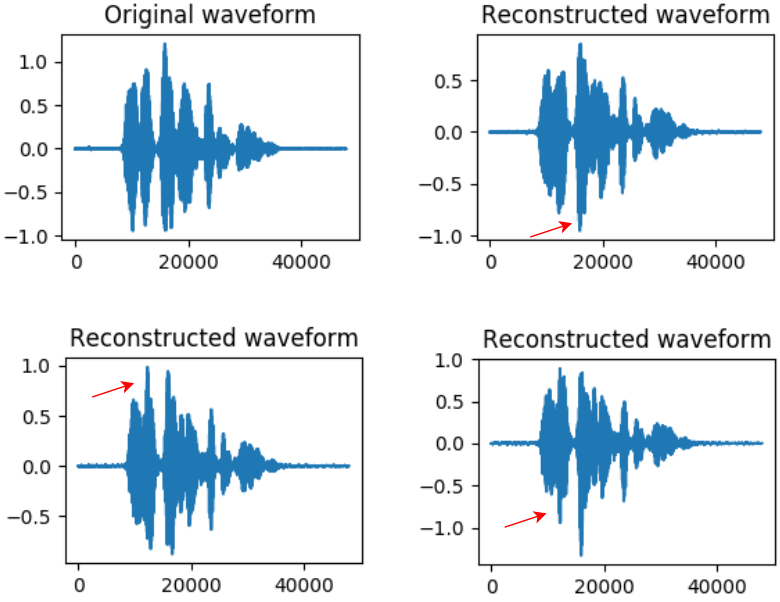}}
\end{minipage}
\caption{\textbf{Diverse predictions} Real ground truth waveform is shown in top-left, remaining are different predictions by the proposed model given the same silent video input.}
\label{diversity}
\vspace{-0.1in}
\end{figure}

To evaluate whether the proposed model captures the human speech's correct distribution, we draw multiple samples from our model given the same frame stream as an input. As shown in Figure~\ref{diversity}, we can generate different plausible speech waveforms by varying z while keeping the input fixed. On the other hand, the existing models that explicitly assume a deterministic one-to-one mapping between the input and output modalities can only generate a single outcome, thus neglecting the multimodality aspect.

\section{Conclusion}
In this work, we present a novel state-of-the-art approach for speech prediction in silent videos. Unlike prior research, the proposed model takes into account the key aspects of human speech, namely that it is multimodal, dynamic, and variable. Our main contribution is an effective stochastic model that can generate virtually infinite high quality audio output sequences for a given silent video, thus capturing the multi-modality of the speech prediction problem. Both qualitative and quantitative results indicate higher quality predictions that are comparable to, or better than, existing approaches.

\bibliographystyle{IEEEbib}
\bibliography{strings,refs}

\end{document}